\documentclass[sigconf]{acmart}

\usepackage{graphicx}

\usepackage{bm}

\AtBeginDocument{%
  \providecommand\BibTeX{{%
    \normalfont B\kern-0.5em{\scshape i\kern-0.25em b}\kern-0.8em\TeX}}}

\setcopyright{acmcopyright}
\copyrightyear{2021}
\acmYear{2021}
\setcopyright{acmlicensed}
\acmConference[MM '21] {Proceedings of the 29th ACM International Conference on Multimedia}{October 20--24, 2021}{Virtual Event, China.}
\acmBooktitle{Proceedings of the 29th ACM Int'l Conference on Multimedia (MM '21), Oct. 20--24, 2021, Virtual Event, China}
\acmPrice{15.00}
\acmISBN{978-1-4503-8651-7/21/10}
\acmDOI{10.1145/3474085.3475172}

\begin{document}
\fancyhead{}
\title{Hierarchical View Predictor: Unsupervised 3D Global Feature Learning through Hierarchical Prediction among Unordered Views}

\author{Zhizhong Han}
\email{h312h@wayne.edu}
\affiliation{%
  \institution{Tsinghua University, Beijing, China and Wayne State University}
  \city{Detroit}
  \state{Michigan}
  \country{USA}
  \postcode{48201}
}
\author{Xiyang Wang}
\email{ssdutwxy@gmail.com}
\affiliation{%
  \institution{School of Software, BNRist, Tsinghua University}
  \city{Beijing}
  \country{P. R. China}
  \postcode{100084}
}

\author{Yu-Shen Liu}
\authornote{Corresponding author. This work was supported by National Key R\&D Program of China (2020YFF0304100), the National Natural Science Foundation of China (62072268), and in part by Tsinghua-Kuaishou Institute of Future Media Data, and NSF (award 1813583).}
\email{liuyushen@tsinghua.edu.cn}
\affiliation{%
  \institution{School of Software, BNRist, Tsinghua University}
  \city{Beijing}
  \country{P. R. China}
  \postcode{100084}
    }

\author{Matthias Zwicker}
\email{zwicker@cs.umd.edu}
\affiliation{%
  \institution{University of Maryland}
  \city{College Park}
  \state{Maryland}
    \country{USA}
  \postcode{20740}
    }

%
%
%
%

\renewcommand{\shortauthors}{Zhizhong Han et al.}

\begin{abstract}
  Unsupervised learning of global features for 3D shape analysis is an important research challenge because it avoids manual effort for supervised information collection.
   In this paper, we propose a view-based deep learning model called \textit{Hierarchical View Predictor} (HVP) to learn 3D shape features from unordered views in an unsupervised manner. To mine highly discriminative information from unordered views, HVP performs a novel hierarchical view prediction over a view pair, and aggregates the knowledge learned from the predictions in all view pairs into a global feature.
   In a view pair, we pose hierarchical view prediction as the task of hierarchically predicting a set of image patches in a current view from its complementary set of patches,
   and in addition, completing the current view and its opposite from any one of the two sets of patches. Hierarchical prediction, in patches to patches, patches to view and view to view, facilitates HVP to effectively learn the structure of 3D shapes from the correlation between patches in the same view and the correlation between a pair of complementary views. In addition, the employed implicit aggregation over all view pairs enables HVP to learn global features from unordered views.
   Our results show that HVP can outperform state-of-the-art methods under large-scale 3D shape benchmarks in shape classification and retrieval.
\end{abstract}




\maketitle

\section{Introduction}
Learning discriminative global features is important for 3D shape analysis tasks such as classification~\cite{Sharma16,WuNIPS2016,Zhizhong2016,Zhizhong2016b,HanCyber17a,YaoqingCVPR2018iccv,PanosCVPR2018ICML,HanTIP18,Zhizhong2018VIP}, retrieval~\cite{Sharma16,WuNIPS2016,Zhizhong2016,Zhizhong2016b,HanCyber17a,YaoqingCVPR2018iccv,PanosCVPR2018ICML,HanTIP18}, correspondence~\cite{MvCNN2017,HanCyber17a,Zhizhong2016,Zhizhong2016b,HanTIP18}, segmentation~\cite{cvprpoint2017,nipspoint17,wenxinacmmm2020}, and reconstruction~\cite{Jiang2019SDFDiffDRcvpr,wenxin_2020_CVPR,wenxin_2021a_CVPR,wenxin_2021b_CVPR,hutaoaaai2020,seqxy2seqzeccv2020paper,Zhizhong2020icml,ZhizhongSketch2020,Zhizhong2021icml,zhizhongiccv2021matching,zhizhongiccv2021completing}. With supervised information~\cite{MvCNN2017,cvprpoint2017,nipspoint17,p2seq18}, recent deep learning based methods have achieved remarkable results under large-scale 3D benchmarks. However, intense manual labeling effort is required to obtain supervised information. In contrast, unsupervised 3D feature learning offers a more promising research challenge that avoids the manual labeling effort.

Several studies have addressed this challenge recently~\cite{YanNIPS2016,Sharma16,WuNIPS2016,Zhizhong2016,Zhizhong2016b,Girdhar16,RezendeEMBJH16,ChoyXGCS16,HanCyber17a,YaoqingCVPR2018iccv,PanosCVPR2018ICML,HanTIP18,Zhizhong2018VIP} by extracting ``supervised'' information in an unsupervised scenario for the training of deep learning models. Extracting self-supervised information is usually achieved by posing different prediction tasks, such as the prediction of a shape from itself by minimizing reconstruction error~\cite{Sharma16,WuNIPS2016,HanCyber17a,YaoqingCVPR2018iccv,PanosCVPR2018ICML} or embedded energy~\cite{Zhizhong2016,Zhizhong2016b}, the prediction of a 3D shape from its context given by 2D views of the shape~\cite{YanNIPS2016,ChoyXGCS16,Zhizhong2018VIP} or local shape features~\cite{HanTIP18}, or the prediction of a shape from views and itself together~\cite{Girdhar16,RezendeEMBJH16}. Among all these methods, multiple sequential views are usually employed to provide a holistic context of 3D shapes, however, unordered views can still not be leveraged to learn.


In this paper, we propose a novel model for 3D shape feature learning using a self-supervised view-based prediction task, which is formulated using unordered views and not restricted to sequential views. As we demonstrate in our results, this leads to highly discriminative 3D shape features and state-of-the-art performance in 3D shape analysis tasks such as classification and retrieval.

A key idea of our deep learning model, called the \textit{Hierarchical View Predictor} (HVP), is that it operates on pairs of opposing views of each shape.
HVP learns to hierarchically make local view predictions in each view pair, i.e., patches to patches, patches to view, and view to view, and then aggregates the knowledge learned from the predictions in all view pairs into global features.
Specifically, unordered views are taken around a 3D shape on a sphere, where a current view and its opposite view form a view pair. Splitting unordered views into multiple view pairs enables HVP to handle the lack of order among views. In each view pair, given a set of patches of the current view, HVP performs hierarchical view prediction by first predicting the complementary set of patches in the current view, and then completing the whole current view and its opposite from any one of the two sets of patches. Hierarchical view prediction aims to learn the structure of 3D shapes from the correlation between patches in the same view and the correlation between appearances in the pair of complementary views. In addition, HVP employs an effective aggregation technique that implicitly aggregates the knowledge learned in the hierarchical view predictions of all pairs of opposing views.
In summary, our significant contributions are as follows:

\begin{figure*}[htb]
  \centering
   \includegraphics[width=\linewidth]{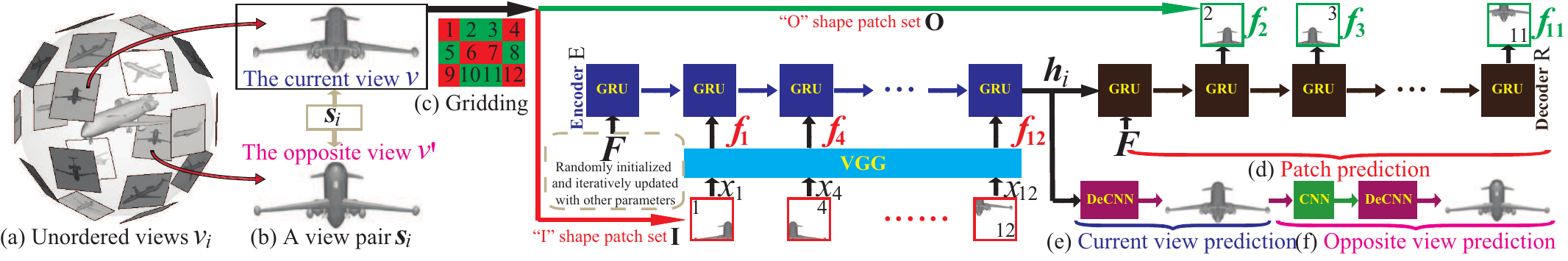}
  %
  %
\caption{\label{fig:Framework}The framework of HVP is demonstrated by learning from a view pair of an plane from (a) to (f).
}
\end{figure*}

\begin{itemize}
\item We propose HVP as a novel deep learning model to perform unsupervised 3D global feature learning through hierarchical view prediction, which leads to state-of-the-art results in classification and retrieval.
\item Hierarchical view prediction enables predictions among unordered views or ordered views, which facilitates HVP to comprehensively understand a 3D shape by hierarchically capturing the correlation between parts in the same view and the correlation between appearances in the pair of complementary views.
\item With simultaneously mining ``supervised'' information inside a view and between two views, HVP eliminates the requirements of learning from dense neighboring view set, which enables HVP to achieve high performance under sparse neighboring view set.
\end{itemize}

\section{Related work}
\noindent\textbf{Supervised 3D feature learning. }With class labels, various deep learning models have been proposed to learn 3D features by capturing the distribution patterns among voxels~\cite{Wu2015,qi2016volumetric,Wangoctree2017}, meshes~\cite{HanTIP18}, points clouds~\cite{cvprpoint2017,nipspoint17,p2seq18,9187572,LiuLRC_2020} and views~\cite{tmmbs2016,Bshi2015,Sfikas17,eccvSinha2017,su15mvcnn,MvCNN2017,Savva2016SHREC,JohnsLD16,chuwang2017,AsakoCVPR2018,parts4features19,9318534,3DViewGraph19,Zhizhong2019seq,3D2SeqViews19}. Among these methods, multi-view based methods perform the best, where pooling is widely used for view aggregation.

\noindent\textbf{Unsupervised 3D feature learning. }To mine ``supervised'' information in unsupervised scenario, deep learning based methods adopted different prediction strategies, such as the prediction of a shape from itself by minimizing reconstruction error~\cite{Sharma16,WuNIPS2016,HanCyber17a,YaoqingCVPR2018iccv,PanosCVPR2018ICML} or embedded energy~\cite{Zhizhong2016,Zhizhong2016b}, the prediction of a shape from context~\cite{YanNIPS2016,ChoyXGCS16,HanTIP18}, or the prediction of a shape from context and itself together~\cite{Girdhar16,RezendeEMBJH16}. These methods employ different kinds of 3D raw representations, such as voxels~\cite{YanNIPS2016,Sharma16,WuNIPS2016,HanCyber17a,Girdhar16,RezendeEMBJH16}, meshes~\cite{Zhizhong2016,Zhizhong2016b,HanTIP18} or point clouds~\cite{DBLP:journals/corr/abs-2107-01886,NIPS2019_SELFREC,YaoqingCVPR2018iccv,PanosCVPR2018ICML,MAPVAE19,Poursaeed20b,gao2020graphter,8885536pc,Hassani_2019_ICCV,l2g2019}, and accordingly, different kinds of context, such as spatial context of virtual words~\cite{HanTIP18} or views~\cite{Girdhar16,ChoyXGCS16,RezendeEMBJH16,YanNIPS2016,Zhizhong2018VIP,gao2021selfsupervised}, are employed. Different from these methods, HVP employs a novel hierarchical view prediction among unordered views to mine more and finer ``supervised'' information.

\noindent\textbf{View synthesis. }Early works teach deep learning models to predict novel views according to input views and transformation parameters~\cite{Dosovitskiy2017}. To generate views with more detail and less geometric distortions, external image sets~\cite{Flynn_2016_CVPR} or geometric constraints~\cite{ZhouTSME16,Ji2017dvm} are further employed. Similarly, the information of multiple past frames is aggregated in video prediction~\cite{WilliamICLR16iccv}. However, these methods cannot aggregate the knowledge learned in each prediction for the discriminability of global features.

\section{Hierarchical view predictor}
The rationale of hierarchical view prediction is to mimic human perception and understanding of 3D shapes. If a human knows a 3D shape, based on observing one part of a view, they can easily imagine the other part of the view, the full view, and even the opposite view. Therefore, HVP mimics this perception by hierarchical view prediction covering patches to patches, patches to view, and view to view, to learn the structure of a 3D shape from the correlation between parts in the same view and the correlation between appearances in the pair of complementary views.

Rather than learning from context by predicting a single patch using CNN in unsupervised image feature learning method~\cite{pathakCVPR16context}, HVP employs an RNN based architecture to predict patch sequence. Since there is only one freely rotated object with the rest of empty background in a rendered view, the context which can be leveraged to learn is much less than in a natural image which contains multiple up-oriented objects. Thus, we want to capture more spatial relationship among parts in a view to remedy the scarce context.

\noindent\textbf{Overview. }The framework of HVP is illustrated in Fig.~\ref{fig:Framework}. By hierarchical view prediction, HVP aims to learn global feature $\bm{F}$ of a 3D shape $m$ from $V$ unordered views $v_i$ ($i\in[1,V]$) taken around $m$ on a sphere. We place the cameras at the 20 vertices of a regular dodecahedron, such that the $V=20$ views are uniformly distributed. The $F$ dimensional feature vector $\bm{F}$ is learned for each shape similarly as in~\cite{Zhizhong2018VIP} via gradient descent together with training the other parameters in HVP, starting from random initialization.

For each shape $m$ in Fig.~\ref{fig:Framework}(a), each view $v_i$ and its opposite view $v_i'$ form a view pair $\bm{s}_i$ in Fig.~\ref{fig:Framework}(b). In $\bm{s}_i$, $v_i$ and $v_i'$ are respectively denoted as $v$ and $v'$ for short. Using a $3 \times 4$ grid, we divide the current view into 12 overlapping patches $x_j$, 
which are further split into two subsets, an ``O'' like shape patch set $\mathbf{O}$ and an ``I'' like shape patch set $\mathbf{I}$, as shown in Fig.~\ref{fig:Framework}(c). Each patch is a square of size $H \times H$.

In a pair $\bm{s}_i$, hierarchical view prediction consists of three prediction tasks in different spaces. Given either one of the two patch sets, HVP first predicts the other patch set in a feature space computed using a VGG19 network as shown in Fig.~\ref{fig:Framework}(d). Then, the current view is predicted in pixel space in Fig.~\ref{fig:Framework}(e). Finally, the opposite view is further predicted based on the predicted current view in Fig.~\ref{fig:Framework}(f).

\begin{figure}[htb]
  \centering
   \includegraphics[width=\linewidth]{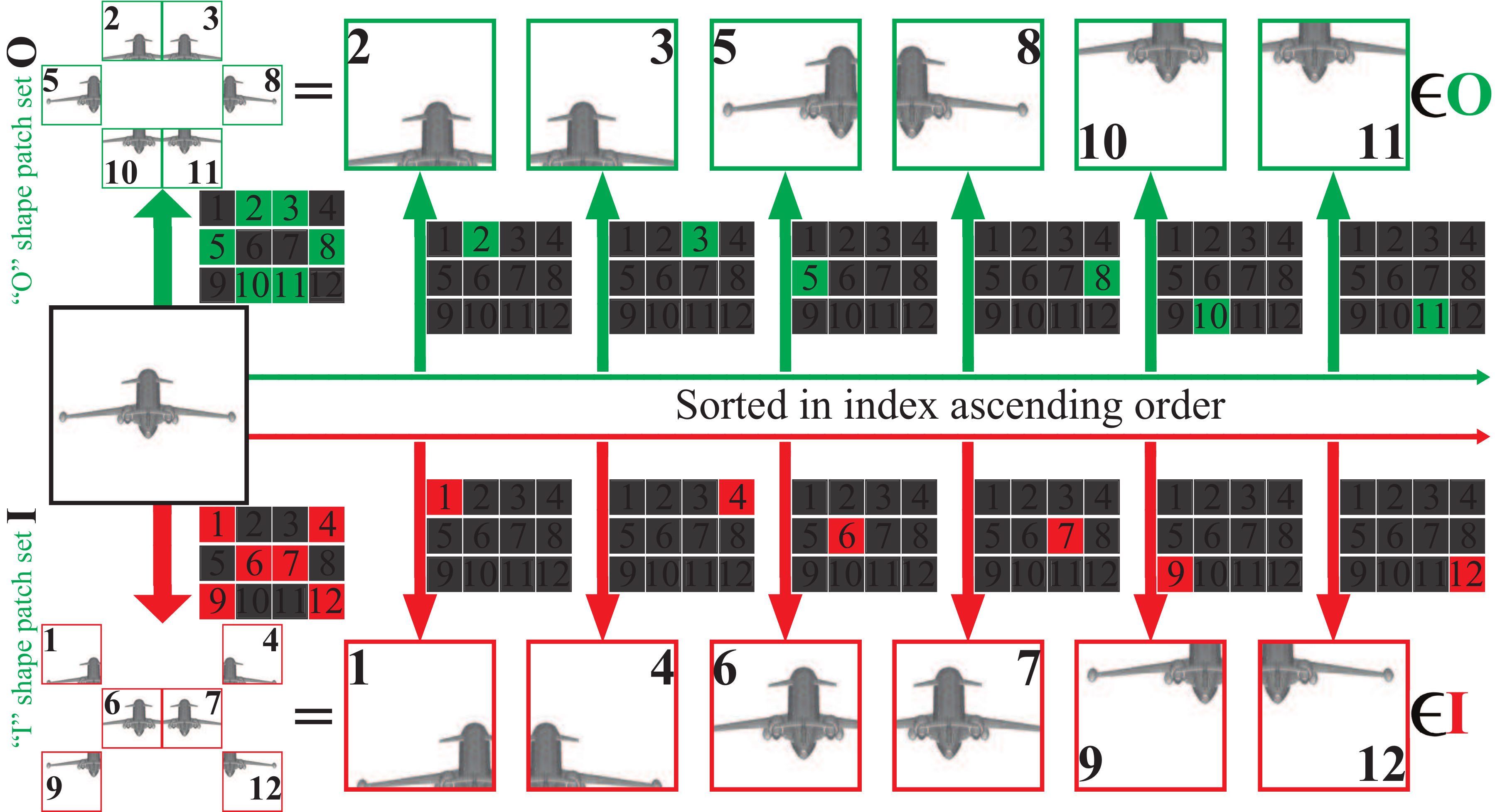}
  %
  %
\caption{\label{fig:Gridding}The gridding procedure splits a view into two patch sets $\mathbf{O}$ and $\mathbf{I}$.
}
\end{figure}

\begin{figure}[htb]
  \centering
   \includegraphics[width=\linewidth]{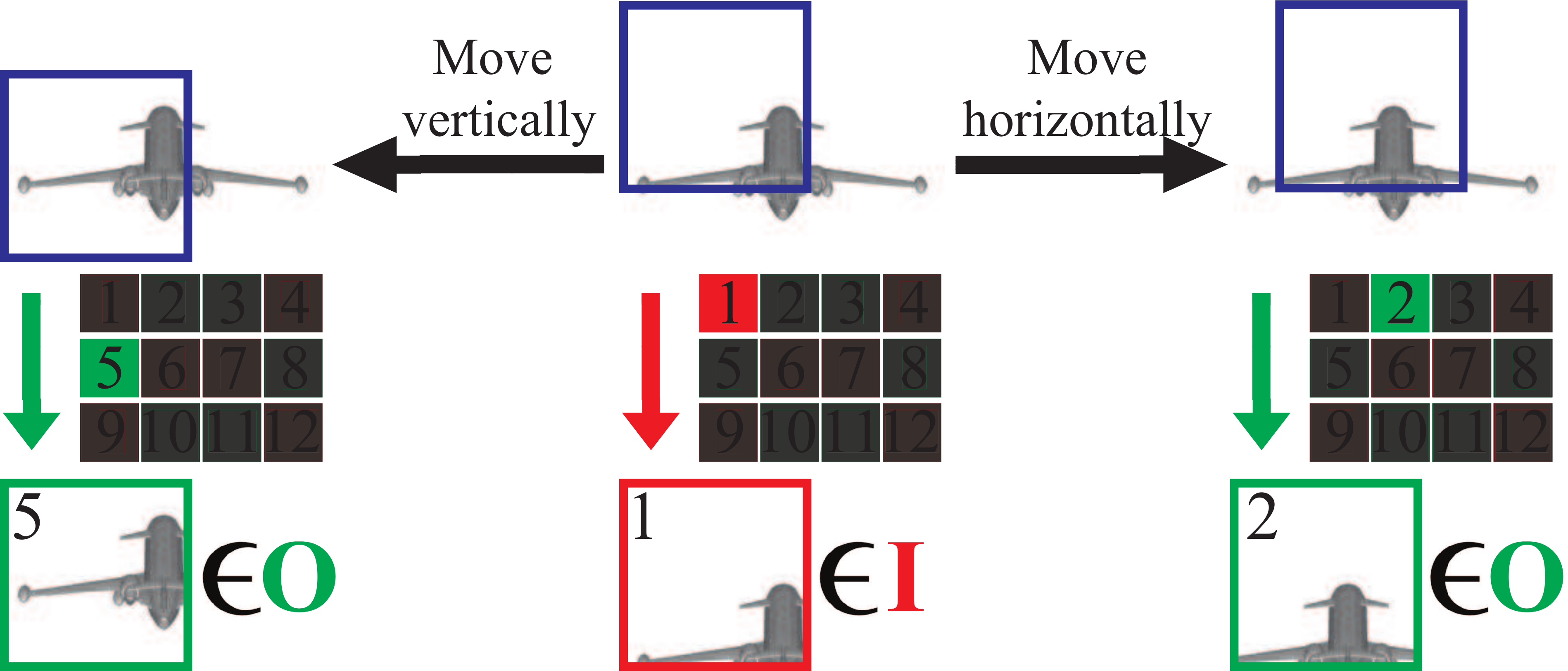}
  %
  %
\caption{\label{fig:Gridding1} The gridding procedure is demonstrated by moving a blue window on a current view in vertical and horizontal direction.}
\end{figure}

\noindent\textbf{Gridding. }In a view pair $\bm{s}_i$, the $224\times 224$ current view $v$ is divided into 12 overlapping patches $x_j$ of size $H\times H$, as demonstrated in Fig.~\ref{fig:Gridding}. The overlapping patches $x_j$ are obtained by uniformly moving a $H\times H$ window across $v$ in vertical and horizontal direction, as demonstrated in Fig.~\ref{fig:Gridding1}. This $3\times 4$ gridding indexes $x_j$ from 1 to 12 in order from left to right and top to bottom. We split the set of 12 patches $x_j$ into an ``O'' like set $\mathbf{O}$ and an ``I'' like set $\mathbf{I}$ of six patches each, i.e., patches $[2,3,5,8,10,11]$ belong $\mathbf{O}$ while the rest patches $[1,4,6,7,9,12]$ belong $\mathbf{I}$, where the patches in $\mathbf{O}$ or $\mathbf{I}$ are sorted into a patch sequence in index ascending order, respectively, as illustrated in Fig.~\ref{fig:Gridding}. We formulate patch prediction as a bidirectional task. That is, both the prediction of $\mathbf{O}$ from $\mathbf{I}$ and the prediction of $\mathbf{I}$ from $\mathbf{O}$ can be conducted using the same network structure (we will take the former for example in the following description).

Our strategy to define and split the patch set is motivated as follows: First, we want to split the patches of the current view into two equally sized subsets to facilitate bidirectional prediction using the same symmetrical structure in HVP. Second, the two patch sets should be distributed over the current view in a similar manner, which eliminates the bias of one directional prediction over the other. Third, each subset should contain a small number of patches to avoid redundancy and increasing computational cost.

\noindent\textbf{Patch prediction. }Patch prediction is performed to capture the correlation between parts in the same view. It predicts the feature $\bm{f}_j$ of patch $x_j$ in set $\mathbf{O}$ (or $\mathbf{I}$) from the features of patches in set $\mathbf{I}$ (or $\mathbf{O}$). According to the indexes established in gridding, we sort all patches in either $\mathbf{O}$ or $\mathbf{I}$ into a sequence in ascending order. This leads us to use a seq2seq model to implement patch prediction as shown in Fig.~\ref{fig:Framework}(d).

We first extract a 4096 dimensional feature $\bm{f}_j$ of each patch $x_j$ by the last fully connected layer of a VGG19 pretrained under ImageNet. Then, we use an encoder RNN $\rm E$ to encode all the $\bm{f}_j$ in $\mathbf{I}$ (denoted as $\bm{f}_{\mathbf{I}}$) with their spatial relationship. We provide the global feature $\bm{F}$ of shape $m$, our learning target, at the first step of the encoder $\rm E$. A key characteristic of our approach is that $\bm{F}$ is shared among all view prediction tasks for each shape $m$. Hence it serves as a knowledge container that keeps incorporating the knowledge derived from each hierarchical view prediction performed on $m$. Different from pooling, which is widely used as an explicit view aggregation, our implicit aggregation enables HVP to mine more and finer information from views of $m$. $\rm E$ encodes $\bm{f}_{\mathbf{I}}$ as a 4096 dimensional hidden state $\bm{h}_i={\rm E}(\bm{f}_{\mathbf{I}})$ of the last step. Finally, based on $\bm{h}_i$, a decoder RNN $\rm R$ is employ to predict the features of patches in $\mathbf{O}$ (denoted as $\bm{f}_{\mathbf{O}}$) with their spatial relationship. Similar to the encoder $\rm E$, we provide the global feature $\bm{F}$ at the first step of decoder $\rm R$, which is regarded as a reference for the following patch feature predictions. For each view, we measure the patch prediction performance of HVP using $L$-2 loss in feature space, denoted as loss $L_{\rm R}$,
\begin{equation}
\label{eq:decoder}
L_{\rm R}=\|{\rm R}({\rm E}(\bm{f}_{\mathbf{I}}))-\bm{f}_{\mathbf{O}}\|^2_2+\|{\rm R}({\rm E}(\bm{f}_{\mathbf{O}}))-\bm{f}_{\mathbf{I}}\|^2_2.
\end{equation}

\noindent\textbf{Current view prediction. }Similar to patch prediction, current view prediction also aims to capture the correlation between parts in the same view, but in pixel space, which forces HVP to understand a 3D shape from the current view in different spaces. HVP predicts the full current view covering both patch sets $\mathbf{I}$ and $\mathbf{O}$ based on the encoding $\bm{h}_i$ of the patch features in set $\mathbf{I}$ or set $\mathbf{O}$. We employ a deconvolutional network $\rm U$ to predict the current view $v$ in pixel space from $\bm{h}_i$, as shown in Fig.~\ref{fig:Framework}(e).

By reshaping the 4096 dimensional $\bm{h}_i$ into 256 feature maps of size $4\times 4$, the deconvolutional network $\rm U$ starts generating $v$ with a resolution of $64\times 64$ through two deconvolutional layers. The two deconvolutional layers employ 128 and 3 kernels, respectively, and each kernel has size $4\times 4$ and a stride of 4, where an ReLu and a tanh are followed in each layer as nonlinear activation functions respectively. For each view, we utilize the $L$-2 loss between the predicted current view ${\rm U}({\rm E}(\bm{f}_{\mathbf{I}}))=\tilde{v}$ (or ${\rm U}({\rm E}(\bm{f}_{\mathbf{O}}))=\tilde{v}$) and the ground truth current view $v$ to measure the current view prediction performance of HVP, denoted as loss $L_{\rm U}$,

\begin{equation}
\label{eq:deconvolution}
L_{\rm U}=\|{\rm U}({\rm E}(\bm{f}_{\mathbf{I}}))-v\|^2_2+\|{\rm U}({\rm E}(\bm{f}_{\mathbf{O}}))-v\|^2_2.
\end{equation}

\noindent\textbf{Opposite view prediction. }We further perform opposite view prediction to capture the correlation between appearances in the pair of complementary views. This helps HVP to bridge one individual view to another and encode their relationship into the global feature. Based on the predicted current view $\tilde{v}$, HVP predicts the opposite view of current view in each view pair $\bm{s}_i$, which is the most challenging task in the three predictions in HVP. This is because the opposite view can only be correctly predicted if the predicted current view $\tilde{v}$ is very close to the ground truth current view $v$, and no other clue is available to use. This challenging criterion pushes HVP to comprehensively learn the intrinsic structure of a 3D shape.

We employ a convolutional network $\rm C$ and the same deconvolutional network $\rm U$ to implement the opposite view prediction, as shown in Fig.~\ref{fig:Framework}(f). The convolutional network $\rm C$ abstracts the $64\times 64$ predicted current view $\tilde{v}$ into a 4096 dimensional feature through three convolutional blocks and one fully connected layer. Each of the three blocks includes 1, 2, and 3 convolutional layers, respectively, which is followed by a maxpool with size of $2\times 2$. All convolutional layers in the three blocks employ kernels with size of $3\times 3$ and a stride of 1, where the ReLu is used as the nolinear activation function. Then, the deconvolutional network $\rm U$ generates the predicted opposite views $\tilde{v}'$ based on the 4096 dimensional feature of $\tilde{v}$. For each view, we also utilize the $L$-2 loss between the predicted opposite view ${\rm U}({\rm C}(\tilde{v}))=\tilde{v}'$ and the ground truth opposite view $v'$ to measure the opposite view prediction performance of HVP, denoted as loss $L_{\rm U}'$,

\begin{equation}
\label{eq:deconvolution}
L_{\rm U}'=\|{\rm U}({\rm C}(\tilde{v}))-v'\|^2_2.
\end{equation}

\noindent\textbf{Objective function. }Finally, in each view pair $\bm{s}_i$, HVP is trained to minimize all the aforementioned losses involved in the three prediction tasks. Therefore, we define the objective function of HVP by combining the three losses as in Eq.~(\ref{eq:HVP}), where the weights $\alpha$ and $\beta$ are used to control the balance among them,
\begin{equation}
\label{eq:HVP}
L=\alpha L_{\rm R}+L_{\rm U}+\beta L_{\rm U}'.
\end{equation}

Note that simultaneously with the other network parameters, we also optimize the learning target $\bm{F}$ by minimizing $L$. We use a standard gradient descent approach by iteratively updating $\bm{F}$ as Eq.~(\ref{eq:update}), where $\varepsilon$ is the learning rate,

\begin{equation}
\label{eq:update}
\bm{F}\gets \bm{F}-\varepsilon\times \partial L / \partial \bm{F}.
\end{equation}

\noindent\textbf{Testing modes. }There are two typical modes of unsupervised learning of features $\bm{F}$ of 3D shapes for testing, which we call the known-test mode and the unknown-test mode. In known-test mode, the test shapes are given with the training shapes at the same time, such that the features of test shapes can be learned with the features of training shapes together. In unknown-test mode, HVP is first pretrained using the set of training shapes only. At test time, we then iteratively learn the features $\mathbf{F}$ of test shapes by minimizing Eq.~(\ref{eq:HVP}) while fixing the other pretrained parameters of HVP.

\section{Experimental results and analysis}
In this section, the performance of HVP is evaluated and analyzed. First we discuss the setup of parameters involved in HVP. These parameters are tuned to demonstrate how they affect the discriminability of learned features in shape classification under ModelNet10~\cite{Wu2015}. Then, some ablation studies are presented to show the effectiveness of some important elements involved in HVP.
Finally, HVP is compared with state-of-the-art methods in shape classification and retrieval under ModelNet10~\cite{Wu2015} and ModelNet40~\cite{Wu2015}. In addition, some generated current views and opposite views are also visualized to justify HVP better. Note that all classification is conducted by training a linear SVM under the global features learned by HVP.

\noindent\textbf{Dataset and evaluations. }The training and testing sets of ModelNet40 consist of 9,843 and 2,468 shapes, respectively. In addition, the training and testing sets of ModelNet10 consist of 3,991 and 908 shapes, respectively.

We employ both average instance accuracy (InsACC) and average class accuracy (ClaACC) to evaluate the classification results. Moreover, we use mAP and precision and recall (PR) curves as metrics in shape retrieval.

\noindent\textbf{Parameter setup. }Initially, the dimension $F$ of global feature $\bm{F}$ is 4096 which is the same as the dimension of $\bm{f}_j$, and the $V=20$ views of all 3D shapes under ModelNet10 are employed to train HVP in known-test mode with a learning rate of $\varepsilon=0.0002$. Both $\alpha$ and $\beta$ are set to 1, which makes the initial values of loss $L_{\rm R}$, $L_{\rm U}$ and $L_{\rm U}'$ comparable to each other, where a normal distribution with mean of 0 and standard deviation of 0.02 is used to initialize the parameters involved in HVP. In addition, the patch width $H$ is 128, and both the prediction of $\mathbf{O}$ from $\mathbf{I}$ and the prediction of $\mathbf{I}$ from $\mathbf{O}$ are performed.

First, we conduct experiments to explore how the learning rate affects the performance of HVP, as shown in Table~\ref{table:balance2}. We employ different learning rates $\varepsilon$ with initial parameters mentioned in the former paragraph. Besides the initial setting, we iteratively use $\varepsilon=\{0.0002,0.0003,0.0004,0.0005,0.0006,0.0007\}$ for training. We achieve the best instance accuracy of $93.61\%$ with $\varepsilon=0.0003$.

\begin{table}
  \caption{The effect of $\varepsilon (\times 0.0001)$ under ModelNet10. $\alpha=1$, $\beta=1$, $H=128$.}
  \label{table:balance2}
  \centering
  \begin{tabular}{c|cccccc}
    \hline
     $\varepsilon$ &2&3&4&5&6&7\\
    \hline		
     InsACC & 91.74 & 92.29 & 92.51 & \textbf{93.61} & 92.73&92.84\\
     ClaACC & 91.48 & 92.05 & 92.15 & \textbf{93.25} & 92.35&92.45 \\
    \hline
  \end{tabular}
\end{table}

Next, we explore the balance weights $\alpha$ and $\beta$. We summarize the results in Table~\ref{table:balance} and Table~\ref{table:balance1}, which shows that the weights are important for the performance of HVP. With $\beta=1$, we explore the effect of $\alpha$ by iteratively setting $\alpha$ to $\{0.25,0.5,1,2,4\}$ in Table~\ref{table:balance}. The instance accuracy increases to a best of $93.61\%$ with $\alpha=1$, and then, decreases gradually.
We observe a similar phenomenon in the exploration of the effect of $\beta$ in Table~\ref{table:balance1}. With $\alpha=1$ obtaining the best result in Table~\ref{table:balance}, we iteratively set $\beta$ to $\{0.25,0.5,1,2,4\}$. The instance accuracy also achieves up to $93.61\%$ with $\beta=1$, and then, decreases a little bit. These results show that both under-fitted and over-fitted patch prediction and opposite view prediction would also affect the discriminability of learned features. In addition, in terms of row averaged accuracies in Table~\ref{table:balance} and Table~\ref{table:balance1}, HVP is affected more by patch prediction than by opposite view prediction, since the row averaged accuracies of $\alpha$ drop more than the ones of $\beta$ from the same highest accuracies.

\begin{table}
  \caption{The effect of $\alpha$ under ModelNet10. $\beta=1$, $H=128$.} 
  \label{table:balance}
  \centering
  \begin{tabular}{c|cccccc}
    \hline
     $\alpha$ &0.25&0.5&1&2&4&Avg\\
    \hline		
     InsACC & 92.18 & 92.84 & \textbf{93.61} & 92.18 & 92.51 & 92.66\\
     ClaACC & 91.85 & 92.75 & \textbf{93.25} & 91.85 & 92.11 & 92.36\\
    \hline
  \end{tabular}
\end{table}

\begin{table}
  \caption{The effect of $\beta$ under ModelNet10. $\alpha=1$, $H=128$.}
  \label{table:balance1}
  \centering
  \begin{tabular}{c|cccccc}
    \hline
     $\beta$ &0.25&0.5&1&2&4&Avg\\
    \hline		
     InsACC & 92.40 & 92.84 & \textbf{93.61} & 93.39 & 92.73 & 92.99\\
     ClaACC & 92.25 & 92.43 & \textbf{93.25} & 92.81 & 92.31 & 92.61\\
    \hline
  \end{tabular}
\end{table}

Finally, we explore how the patch size $H$ affects the performance of HVP, as shown in Table~\ref{table:balance4}. Beside the patch size $H=128$ involved in the former experiments, we also iteratively employ patches with size of $\{160,180,200\}$. Based on a view with size of 224 in our experiments, the increasing patch size $H$ makes the results achieve up to $94.16\%$ with $H=160$, while decreasing gradually for even larger patches. These results show that both a lack of semantic information in small patches and too much redundant information among big overlapping patches are not helpful to increase the discriminability of learned features. Since smaller sizes make the features of the patches meaningless, while big sizes provide too much redundant information, such that HVP can easily solve the patch prediction, without needing to store information in the global feature $\mathbf{F}$ that is shared among all predictions for each shape.

\begin{table}
  \caption{The effect of patch size $H$ under ModelNet10. $\alpha=1$, $\beta=1$.}
  \label{table:balance4}
  \centering
  \begin{tabular}{c|cccc}
    \hline
      Size & 128 & 160 & 180 & 200 \\ 
    \hline		
     InsACC & 93.61 & \textbf{94.16} & 93.94 & 93.73 \\
     ClaACC & 93.25 & \textbf{93.80} & 93.60 & 93.36 \\
    \hline
  \end{tabular}
\end{table}


\noindent\textbf{Ablation studies. }Based on the former experiments, we further explore the contribution of each prediction involved in hierarchical view prediction, as highlighted by results in Table~\ref{table:balance5}. We first train HVP only using the current view prediction loss ($L_{\rm U}$), then, we incrementally add opposite view prediction loss ($\beta L_{\rm U}'$) or patch prediction loss ($\alpha L_{\rm R}$), finally, we compare these results with our best results employing all the three losses ($L_{\rm U}+\beta L_{\rm U}'+\alpha L_{\rm R}$).

\begin{figure}[htb]
  \centering
   \includegraphics[width=\linewidth]{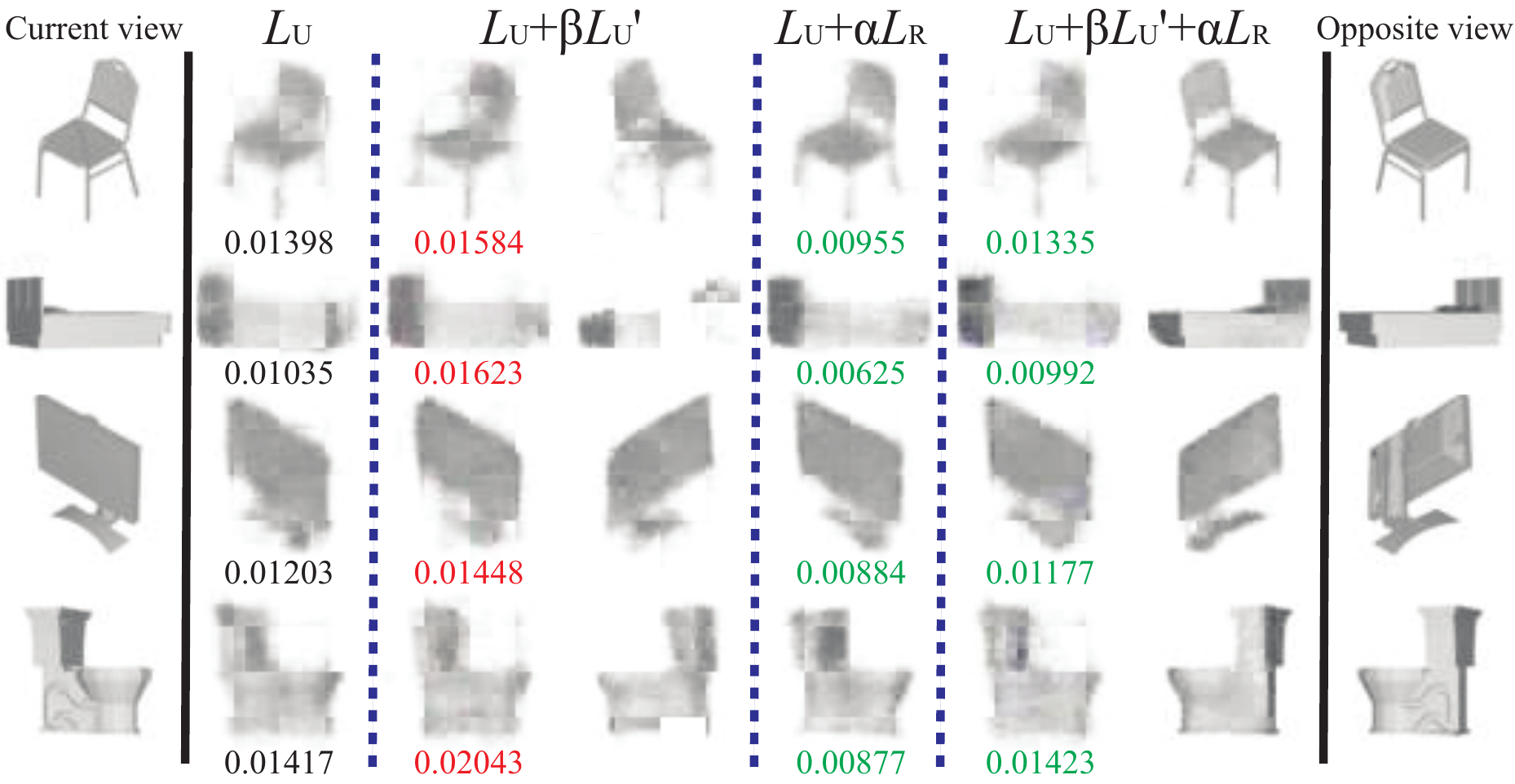}
  %
  %
\caption{\label{fig:AblationViews} The effect of different losses on the generated current view and opposite view in a view pair under ModelNet10. The distance between each generated current view and ground truth is also shown under each generated current view, where the red color means the distance is larger than the distance in the first column in the same row while the green color means it becomes smaller.}
\end{figure}

Compared to the results of ``$L_{\rm U}$'', each incrementally added loss can improve the performance of HVP in terms of both averaged instance accuracy and averaged class accuracy. In addition, the patch prediction loss can help HVP improve more than the opposite view prediction loss. To better visualize the effect of these losses on HVP, we show the generated current view or generated opposite view involved in the experiments in Table~\ref{table:balance5}. In Fig.~\ref{fig:AblationViews}, the generated views and their distances to the ground truth are shown. The added opposite view prediction loss $\beta L_{\rm U}'$ can degenerate the generated current view compared to the one with only $L_{\rm U}$ or the one with $L_{\rm U}+\alpha L_{\rm R}$. While the added patch prediction loss $\alpha L_{\rm R}$ can improve the generated current view or the generated opposite view, as shown by the comparison between $L_{\rm U}$ and $L_{\rm U}+\alpha L_{\rm R}$ and the comparison between $L_{\rm U}+\alpha L_{\rm R}$ and $L_{\rm U}+\beta L_{\rm U}'+\alpha L_{\rm R}$.

\begin{table}
  \caption{The contribution of each prediction involved in hierarchical view prediction under ModelNet10. $\alpha=1$, $\beta=1$,$H=160$.} 
  \label{table:balance5}
  \centering
  \begin{tabular}{c|cccc}
    \hline
      Loss & $L_{\rm U}$ & $L_{\rm U}+\beta L_{\rm U}'$ & $L_{\rm U}+\alpha L_{\rm R}$ & $L$ \\
    \hline		
     InsACC & 86.78  & 88.66 & 90.31 & \textbf{94.16} \\
     ClaACC & 86.41  & 87.80 & 90.10 & \textbf{93.80} \\
    \hline
  \end{tabular}
\end{table}

Then, we highlight the effect of prediction direction in Table~\ref{table:balance6}. In all the former experiments, we employ the bidirectional prediction
This could provide more data to train HVP thanks to the symmetrical structure of HVP. In the following experiments, we train HVP using different kinds of single directional data respectively, such as from $\mathbf{O}$ to $\mathbf{I}$ (shown as $\mathbf{O}2\mathbf{I}$), from $\mathbf{I}$ to $\mathbf{O}$ (shown as $\mathbf{I}2\mathbf{O}$), and randomly selected single direction for hierarchical view prediction on each view (shown as $\mathbf{O}2\mathbf{I}$ or $\mathbf{I}2\mathbf{O}$). As shown by these results, bidirectional prediction could learn better features than single direction prediction.

\begin{table}
  \caption{The effect of prediction direction under ModelNet10. $\alpha=1$, $\beta=1$, $H=160$.} 
  \label{table:balance6}
  \centering
  \begin{tabular}{c|cccc}
    \hline
      Direction & $\mathbf{O}2\mathbf{I}$ & $\mathbf{I}2\mathbf{O}$ & $\mathbf{O}2\mathbf{I}$ or $\mathbf{I}2\mathbf{O}$ & $\mathbf{O}2\mathbf{I}$ and $\mathbf{I}2\mathbf{O}$\\
    \hline		
     InsACC & 93.83 & 93.61 & 93.72 & \textbf{94.16} \\
     ClaACC & 93.38 & 93.16 & 93.25 & \textbf{93.80} \\
    \hline
  \end{tabular}
\end{table}

Finally, we justify our view aggregation method in Table~\ref{table:balance7}. We first train HVP without $\bm{F}$. To obtain a global shape feature without $\bm{F}$, we use pooling on the features $\bm{h}_i$ for all view pairs $\bm{s}_i$ for each shape (recall that we use the $\bm{h}_i$ to solve the view prediction tasks as shown in Fig.~\ref{fig:Framework}).
The results using mean and max pooling without $\bm{F}$ in the first column in Table~\ref{table:balance7} exhibit a large drop in performance compared to our approach. To better understand this, we perform a second experiment where we train HVP with $\bm{F}$ as described previously, but we again use the mean- or max-pooled features $\bm{h}_i$ as the global shape feature. The results in the second column in Table~\ref{table:balance7} (first and second row) show that using $\bm{F}$ improves the performance of the mean- and max-pooled features $\bm{h}_i$. However, our approach that uses $\bm{F}$ itself as the shape feature (third row) achieves even better performance, indicating that HVP's implicit pooling is superior to explicit schemes such as mean- or max-pooling. Intuitively, this is because max pooling can lose some information in each view pair, while mean pooling weights all view pairs equally. Hence it fails to give more weight to certain highly distinct view pairs.

\begin{table}
  \caption{The effect of view aggregation under ModelNet10. $\alpha=1$, $\beta=1$, $H=160$.}
  \label{table:balance7}
  \centering
  \begin{tabular}{c|cc|cc}
    \hline
      &\multicolumn{2}{|c|}{Without $\bm{F}$}&\multicolumn{2}{|c}{With $\bm{F}$}\\
      \hline
      Methods & InsACC & ClaACC & InsACC & ClaACC\\
      \hline
      MeanPool & 89.65 & 87.96 & 90.75 & 90.58\\
      MaxPool & 90.31 & 89.84 & 91.19 & 90.92\\
      \hline
      Our & - & - & \textbf{94.16} & \textbf{93.80} \\
    \hline
  \end{tabular}
\end{table}


\begin{table}
  \caption{Classification comparison under MN10.}
  \label{table:comparison1}
  \centering
  \begin{tabular}{cccc}
    \hline
    Methods & Supervised  & Instance$\%$ & Class$\%$ \\
    \hline
    ORION\cite{SZB17a} & Yes &-& 93.80 \\
    3DDescriptorNet\cite{JianwenCVPR2018} & Yes &-& 92.40 \\
    Pairwise\cite{JohnsLD16} & Yes & -&92.80\\
    GIFT\cite{tmmbs2016} & Yes & -&91.50 \\
    VoxNet\cite{Maturana15} & Yes & -&92.00 \\
    VRN\cite{Brocknips2016}& Yes & 93.80 & - \\
    PANORAMA\cite{Sfikas17} & Yes & -&91.12 \\
    \hline
    LFD\cite{Chen03} & No &  79.90 & -  \\
    Vconv-DAE\cite{Sharma16} & No &-& 80.50 \\
    3DGAN\cite{WuNIPS2016} & No & -&91.00 \\
    VSL\cite{liu2018learning} & No & 91.00 & - \\
    NSampler\cite{Edoardo19} & no & 88.70 & 95.30 \\
    LGAN\cite{PanosCVPR2018ICML} & No & 95.30& - \\
    LGAN\cite{PanosCVPR2018ICML}(MN10) & No & 92.18& - \\
    FNet\cite{YaoqingCVPR2018iccv} & No & 94.40& -\\
    FNet\cite{YaoqingCVPR2018iccv}(MN10) & No & 91.85& -\\
    VIPGAN\cite{Zhizhong2018VIP}& No &94.05&93.71\\
    \hline
    Our & No & \textbf{94.16} &\textbf{93.80} \\ 
    Our(MN40) & No & \textbf{92.18} &\textbf{91.57} \\
    \hline
  \end{tabular}
\end{table}

\noindent\textbf{Classification. }We compare HVP with the state-of-the-art methods in classification under ModelNet10 and ModelNet40 in Table~\ref{table:comparison1} and Table~\ref{table:comparison}, respectively. The parameters under ModelNet40 are the same ones with our best results under ModelNet10 in Table~\ref{table:balance7}. Under ModelNet10, HVP outperforms all its unsupervised competitors under ModelNet10, as shown by ``Our'', which is also the best result compared to eight top ranked supervised methods. Under ModelNet40, HVP achieves the state-of-the-art results among all the unsupervised and supervised competitors. For fair comparison, the result of VRN~\cite{Brocknips2016} is presented without ensemble learning. The result of RotationNet~\cite{AsakoCVPR2018} is presented with views taken by the default camera system orientation that is identical to the others.
Although the results of LGAN, FNet and NSampler are better than our results under ModelNet10, it is inconclusive whether they are better than ours. This is because these methods are trained under a version of ShapeNet55 that contains more than 57,000 3D shapes, including a number of 3D point clouds. However, there are only 51,679 3D shapes from ShapeNet55 that are available for public download. Therefore, we cannot use the same amount of training data to train HVP to compare with them. To perform fair comparison with ``Our'', we use the codes of LGAN and FNet to conduct experiments only using shapes in ModelNet, as shown by ``LGAN()'' and ``FNet()'', which employs the same training data as ours. Our performing results show that our method is superior to these methods.

With the ability of mining correlation inside each view, HVP does not rely on dense neighboring views to learn. To justify this, we train HVP using two views of each 3D shape from ModelNet40. Although VIPGAN achieves the best results with 12 views under ModelNet40, HVP works much better than VIPGAN under sparse neighboring view set, as shown by the comparison between ``Our(Two)'' and ``VIPGAN(Two)'' in Table~\ref{table:comparison}.

Moreover, we evaluate HVP in unknown-test mode by learning features of ModelNet10 using parameters pretrained under ModelNet40 (``Our'' in Table~\ref{table:comparison}). As shown by ``Our(MN40)'' in Table~\ref{table:comparison1}, HVP can still produce good results. These results show that HVP is with remarkable transfer learning ability based on comprehensive 3D shapes understanding, which is benefited by mining correlation inside a view and between complementary views.

\begin{table}
  \caption{Classification comparison under MN40.}
  \label{table:comparison}
  \centering
  \begin{tabular}{cccc}
    \hline
    Methods & Supervised  & Instance$\%$ & Class$\%$ \\
    \hline
    MVCNN\cite{su16mvcnn} & Yes & 92.0 & 89.7\\
    MVCNN-Sphere\cite{su16mvcnn} & Yes & 89.5 & 86.6\\
    Pairwise\cite{JohnsLD16}& Yes & -&90.70\\
    GIFT\cite{tmmbs2016} & Yes & - &89.50\\
    PointNet++\cite{nipspoint17} & Yes & 91.90&-\\
    VRN\cite{Brocknips2016} & Yes & 91.33&- \\
    RotationNet\cite{AsakoCVPR2018} & Yes & 92.37&-\\
    PANORAMA\cite{Sfikas17} & Yes& - & 90.70\\
    \hline
    T-L Network\cite{Girdhar16} & No & - & 74.40\\
    Vconv-DAE\cite{Sharma16} & No & - & 75.50\\
    3DGAN\cite{WuNIPS2016} & No &- & 83.30\\
    VSL\cite{liu2018learning} & No & 84.50 & - \\
    LGAN\cite{PanosCVPR2018ICML} & No & 85.70&-\\
    LGAN\cite{PanosCVPR2018ICML}(MN40) & No & 87.27&-\\
    FNet\cite{YaoqingCVPR2018iccv} & No& 88.40 &-\\
    FNet\cite{YaoqingCVPR2018iccv}(MN40) & No& 84.36 &-\\
    NSampler\cite{Edoardo19} & no & 88.70 & - \\
    MRTNet\cite{mrt18} & No& 86.40& -\\
    3DCapsule\cite{YonghengCVPR2019} & No & 88.90& -\\
    PointGrow\cite{YongbinCVPR2019} & No & 85.80 & -\\
    PCGAN\cite{ChunLiangCVPR2019}& No & 87.80 & -\\
    MAPVAE\cite{MAPVAE19}& No & 90.15 & -\\
    OrientNet\cite{Poursaeed20b}& No & 90.75 & -\\
    VIPGAN\cite{Zhizhong2018VIP}& No &91.98&-\\
    VIPGAN(Two)\cite{Zhizhong2018VIP}& No &4.05&2.50\\
    \hline
    Our & No & \textbf{90.72} &\textbf{87.95}\\ 
    Our(Two) & No & \textbf{88.33} &\textbf{83.53}\\ 
    \hline
  \end{tabular}
\end{table}

\begin{figure}[!htb]
  \centering
   \includegraphics[width=\linewidth]{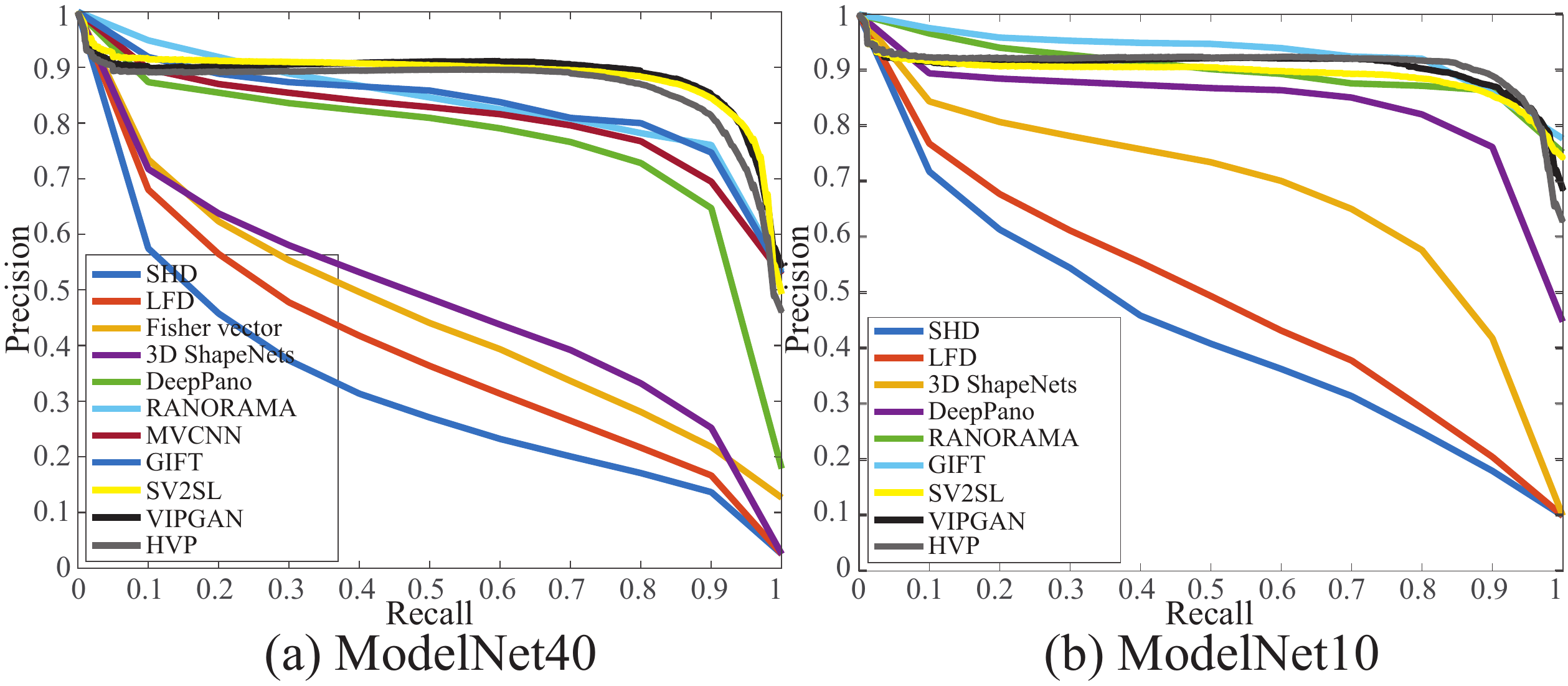}
  %
  %
\caption{\label{fig:PR}The PR curve comparison under ModelNet40 and ModelNet10.}
\end{figure}

\noindent\textbf{Retrieval. }We further evaluate HVP in shape retrieval under the ModelNet40 and ModelNet10 by comparing with the state-of-the-art methods in Table~\ref{table:retrieval}. These experiments are conducted under the test set, where each 3D shape is used as a query to retrieve from the rest of the shapes, and the retrieval performance is evaluated by mAP. In addition, we employ the same parameters with our best classification results in Table~\ref{table:comparison} and Table~\ref{table:comparison1} to extract the global features for the retrieval experiments under ModelNet40 and ModelNet10, respectively. As shown in Table~\ref{table:retrieval}, our results outperform all the compared results under ModelNet10 and achieve state-of-the-art under ModelNet40. In addition, the available PR curves under ModelNet40 and ModelNet10 are also compared in Fig.~\ref{fig:PR}, which also demonstrates our outperforming results in shape retrieval.

\begin{table}[h]
\centering
\caption{The comparison of retrieval in terms of mAP under ModelNet40 and ModelNet10.}  
    \begin{tabular}{|c|c|c|c|}  
     \hline
       Methods & Range & MN40 & MN10 \\  
     \hline
       SHD~\cite{Kazhdan03} & Test-Test & 33.26 & 44.05 \\
       LFD~\cite{Chen03} & Test-Test & 40.91 & 49.82 \\
       3DShapeNets~\cite{Wu2015} & Test-Test & 49.23 & 68.26 \\
       GeomImage~\cite{eccvSinha2017} & Test-Test & 51.30 & 74.90 \\
       DeepPano~\cite{Bshi2015} & Test-Test & 76.81 & 84.18 \\
       MVCNN~\cite{su15mvcnn} & Test-Test & 79.50 & - \\
       PANORAMA~\cite{Sfikas17} & Test-Test & 83.45 & 87.39 \\
       GIFT~\cite{tmmbs2016} & Random & 81.94 & 91.12\\
	   Triplet~\cite{Xinweicvpr18} & Test-Test & 88.00 & - \\
       SliceVoxel~\cite{Miyagi18} & Test-Test & 77.48 & 85.34 \\
       SV2SL~\cite{Zhizhong2018seq} & Test-Test & 89.00 & 89.55\\
       VIPGAN~\cite{Zhizhong2018VIP}& Test-Test & 89.23 & 90.69\\
       Serial~\cite{Xu_2019_ICCV}&Test-Test&87.05&-\\
     \hline
     Ours& Test-Test & \textbf{87.13} & \textbf{91.19}\\
     \hline
   \end{tabular}
   \label{table:retrieval}
\end{table}

\begin{figure*}[!htb]
  \centering
   \includegraphics[width=\linewidth]{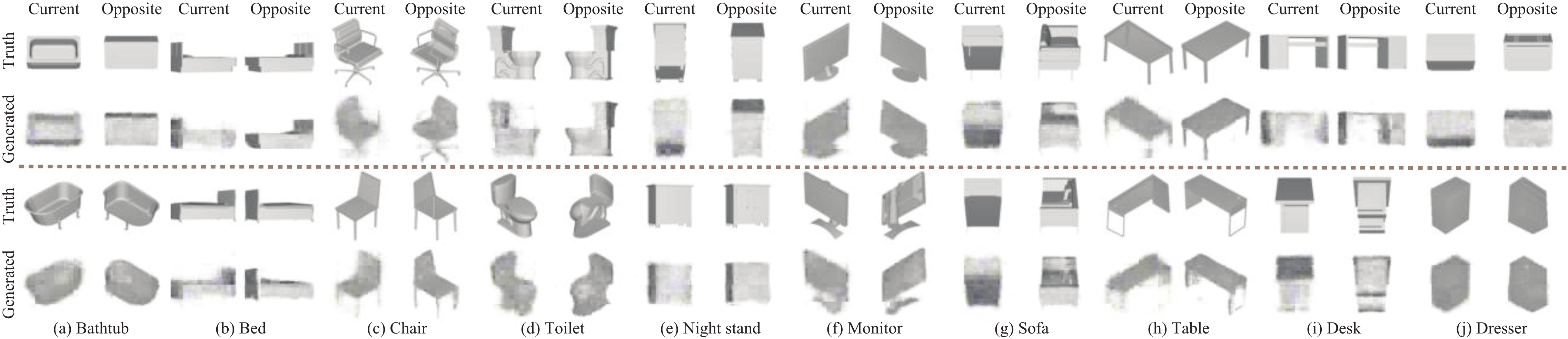}
  %
  %
\caption{\label{fig:Visualization}The generated current view and generated opposite view of a shape are visualized.
}
\end{figure*}

\begin{figure}[htb]
  \centering
   \includegraphics[width=\linewidth]{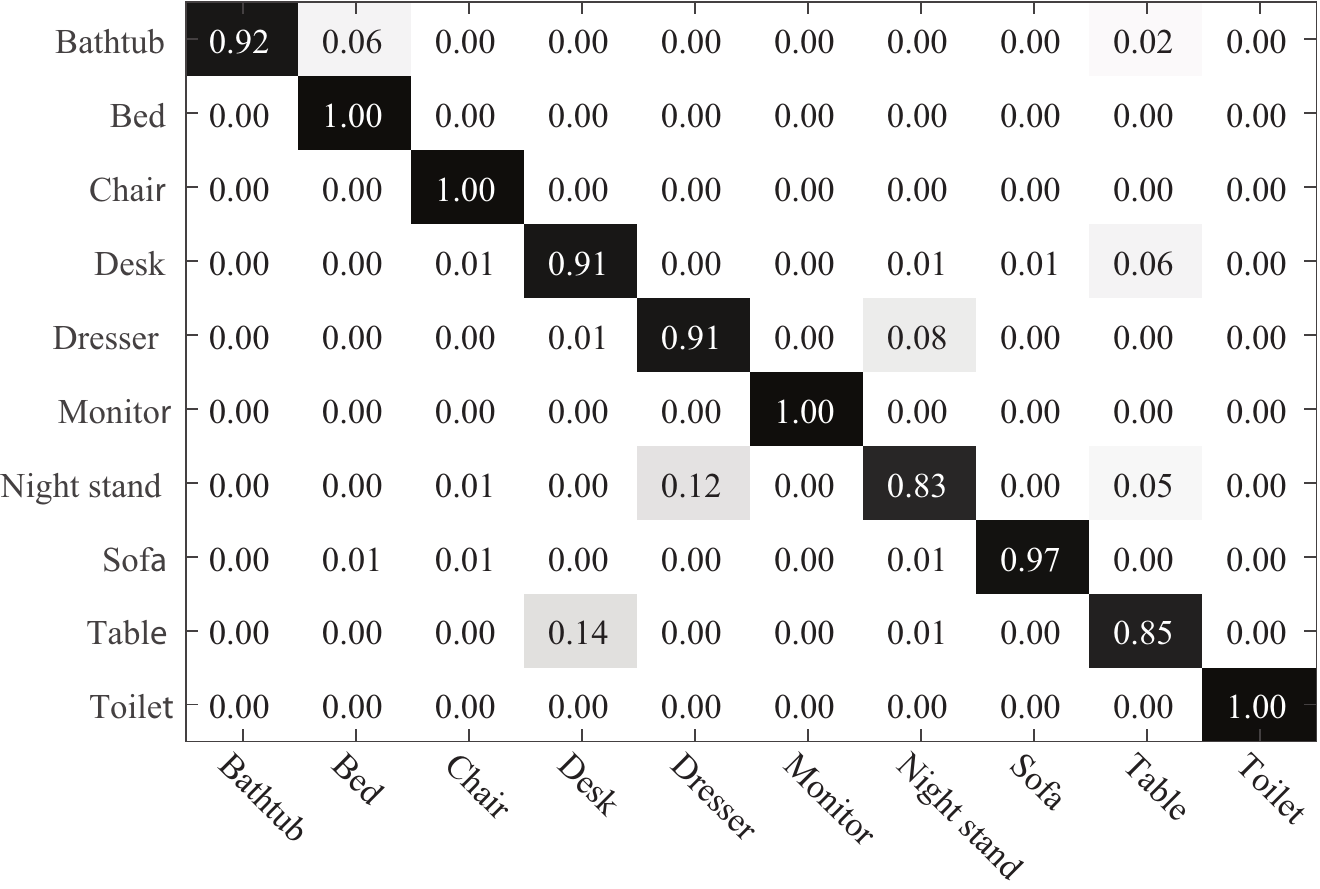}
  %
  %
\caption{\label{fig:MN10Confusion} The confusion matrix of our results in 3D shape classification under ModelNet10.
}
\end{figure}
\begin{figure}[htb]
  \centering
   \includegraphics[width=\linewidth]{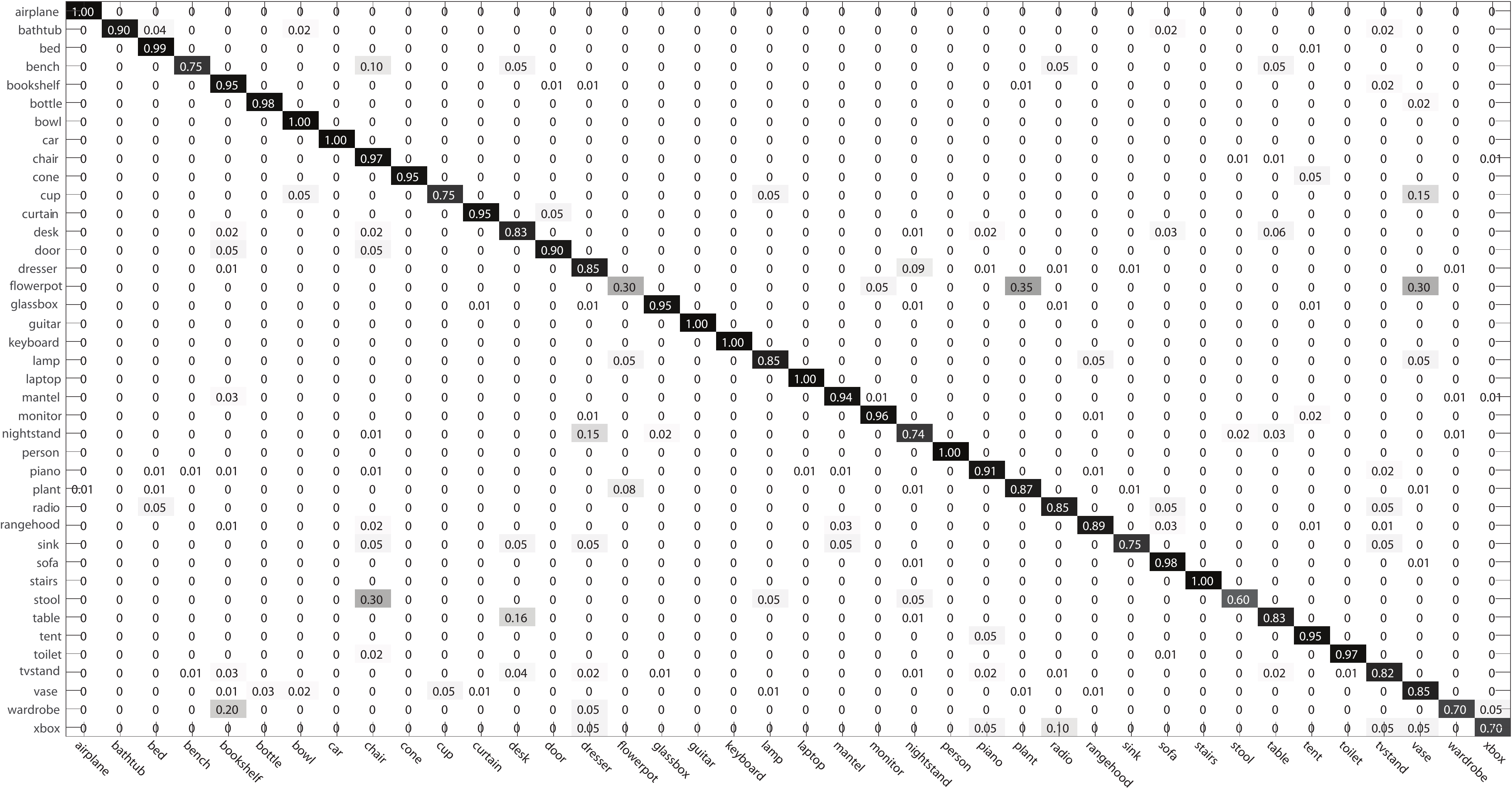}
  %
  %
\caption{\label{fig:MN40Confusion} The confusion matrix of our results in 3D shape classification under ModelNet40.
}
\end{figure}

\noindent\textbf{Visualization. }For our classification results, we show the generated current views and the generated opposite views of all ten shape classes under ModelNet10, where two shapes are involved in each shape class, as shown in Fig.~\ref{fig:Visualization}. These results show that HVP can generate plausible views based on the comprehensive understanding of 3D shapes. Another interesting observation is that the opposite view can be generated better than the current view. In addition, we show the confusion matrix of our classification results under ModelNet10 and ModelNet40 in Fig.~\ref{fig:MN10Confusion} and Fig.~\ref{fig:MN40Confusion}, respectively. In each confusion matrix, an element in the diagonal line means the percentage of how many 3D shapes are correctly classified, while other elements in the same row means the percentage of 3D shapes wrongly classified into other shape classes. The large diagonal elements in each confusion matrix show that HVP is able to learn highly discriminative features for 3D shapes, which facilitates HVP to achieve high performance in classifying large-scale 3D shapes.

\begin{figure}[!htb]
  \centering
   \includegraphics[width=\linewidth]{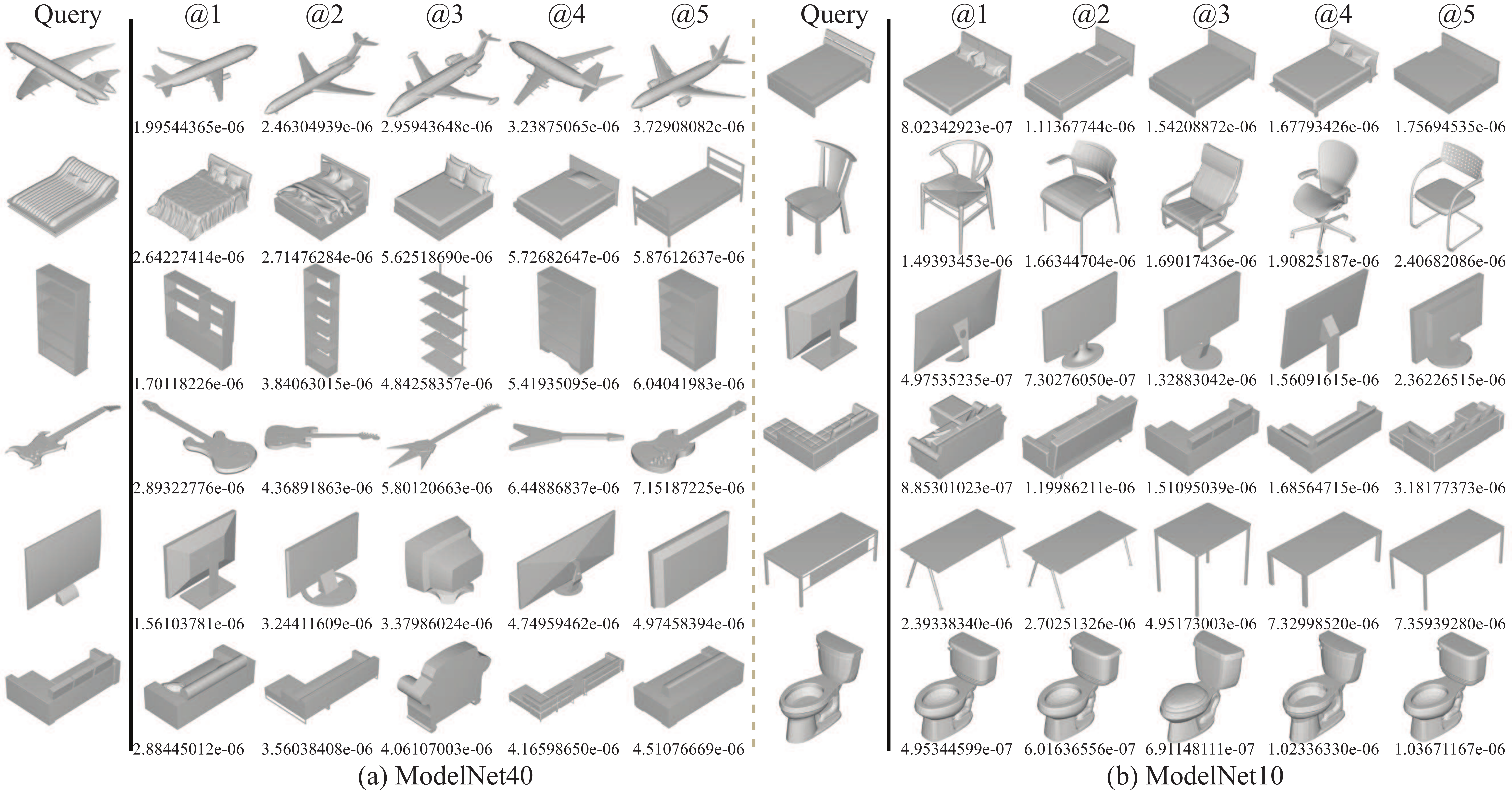}
  %
  %
\caption{\label{fig:RetrievalVisualization}The top 5 retrieved 3D shapes for some queries from (a) ModelNet40 and (b) ModelNet10 are demonstrated. The distance between the query and each retrieved shape is also shown.
}
\end{figure}

For our retrieval results, we show the top 5 retrieved 3D shapes for some queries from ModelNet10 and ModelNet40 in Fig.~\ref{fig:RetrievalVisualization}. According to the distance between the query and each retrieved shape (shown under each retrieved shape), we find HVP is capable of learning features to distinguish 3D shapes in detail, which enables retrieving 3D shapes with similar structures.

\section{Conclusions}

We proposed HVP for unsupervised 3D global feature learning from unordered views of 3D shapes. By implementing a novel hierarchical view prediction, HVP successfully mines highly discriminative information among unordered views in an unsupervised manner. Our results show that HVP effectively learns to hierarchically make patch predictions, current view prediction and opposite view prediction in each view pair, and then, comprehensively aggregates the knowledge learned from the predictions
in all view pairs into global features. HVP can not only learn from both unordered and ordered view set, but also work well under sparse neighboring view sets, which eliminates the requirement of mining ``supervised'' information from dense neighboring views. Our results show that HVP outperforms its unsupervised counterparts, as well as some top ranked supervised methods under large-scale benchmarks in shape classification and retrieval.


\bibliographystyle{ACM-Reference-Format}
\bibliography{paper}

\end{document}